\documentclass[conference]{IEEEtran}
\IEEEoverridecommandlockouts
\usepackage{cite}
\usepackage{amsmath,amssymb,amsfonts}
\usepackage{algorithm}
\usepackage{graphicx}
\usepackage{textcomp}
\usepackage{xcolor}
\usepackage{booktabs}
\usepackage{multirow}
\usepackage{amsthm}
\usepackage{url}
\usepackage{array}
\usepackage[utf8]{inputenc}
\usepackage[T1]{fontenc}

\def\BibTeX{{\rm B\kern-.05em{\sc i\kern-.025em b}\kern-.08em
    T\kern-.1667em\lower.7ex\hbox{E}\kern-.125emX}}
\begin{document}

\title{Decoupled Hierarchical Reinforcement Learning with State Abstraction for Discrete Grids
\\}
\author{
\IEEEauthorblockN{
Qingyu Xiao,
Yuanlin Chang,
Youtian Du\IEEEauthorrefmark{1}\thanks{\IEEEauthorrefmark{1} Corresponding author: duyt@xjtu.edu.cn}\thanks{
This work was supported by the National Key Research and Development Program of China under Grant 2021YFB2400800.}
}
\\
\IEEEauthorblockA{
School of Automation Science and Engineering, Xi'an Jiaotong University, Xi'an, China
}
}

\maketitle

\begin{abstract}
Effective agent exploration remains a core challenge in reinforcement learning (RL) for complex discrete state-space environments,
particularly under partial observability. This paper presents a decoupled hierarchical RL framework
integrating state abstraction (DcHRL-SA) to address this issue. The proposed method employs a dual-level
architecture, consisting of a high-level RL-based actor and a low-level rule-based policy, to promote effective exploration.
Additionally, state abstraction method is incorporated to cluster discrete states, effectively lowering state dimensionality.
Experiments conducted in two discrete customized grid environments demonstrate that the proposed approach consistently outperforms PPO in terms of exploration efficiency, convergence speed, cumulative reward, and policy stability.
These results demonstrate a practical approach for integrating decoupled hierarchical policies and state abstraction in discrete grids with large-scale exploration space. Code will be available at \url{https://github.com/XQY169/DcHRL-SA}.
\end{abstract}

\section{Introduction}
Reinforcement learning (RL) learns optimal policies through trial-and-error interactions, demonstrating strong adaptability and long-term optimization capabilities, with wide applications in robotics, path planning, and intelligent systems \cite{b1,b2,b3}.
Despite these advances, practical RL applications often suffer from large-scale exploration spaces and partial observability, severely restricting the efficiency of exploration and the overall performance of the policy \cite{b4,b5}.
In particular, discrete state-space environments and partially observable markov decision processes (POMDPs)
present additional difficulties for conventional RL algorithms. In discrete state spaces, the lack of natural similarity metrics hinders policy generalization to unvisited states. Sparse rewards further exacerbate the difficulty in extracting effective learning signals. For POMDPs, partial observability leads to compromised decisions and challenges the efficiency and stability of standard Markov-based RL algorithms.

Multiple strategies have been proposed to address these challenges, including hierarchical reinforcement learning (HRL)
and state abstraction. HRL aims to decompose complex tasks into multiple subtasks, where high-level and low-level policies are responsible
for sub-goal planning and action execution, respectively, thus significantly improving exploration efficiency and learning performance
\cite{b6,b7}. However, most existing methods rely on two coupled RL structures, making the engineering implementation
complex. To address this, some frameworks combining RL with optimization-based controllers have been proposed.
For example, Watanabe et al. \cite{b8} presented a hierarchical strategy that combines RL for high-level decision making with proportional integral derivative (PID) controllers for low-level motion execution, achieving sample-efficient locomotion adaptation
for quadruped robots in complex environments. Similarly, Zhang and Ramirez-Amaro \cite{b9} proposed a framework that integrates symbolic planning with low-level
RL, where the planner generates task sequences for high-level strategies while RL independently learns action
sequences for each subtask. This approach improves efficiency and success rates in complex robotic manipulation tasks compared to
coupled hierarchical methods. Nevertheless, such hierarchical frameworks often require manually specified high-level goals.
Overly coarse goals (e.g. the final destination) compromise the purpose of hierarchical RL, while overly fine-grained goals demand extensive human prior knowledge.

State abstraction reduces exploration complexity by clustering the state space while preserving critical information\cite{b10,b11,b12}.
Deep markov decision process state abstraction (DeepMDP)\cite{b11} strictly preserves the reward and transition equivalence within the abstracted state representation.
This approach has demonstrated promising results in continuous-state RL tasks.
In POMDPs, agents must learn policies based on historical observations or latent state estimations.
By introducing belief distributions, a POMDP can theoretically be transformed into a Markov decision process (MDP) in the belief space\cite{b13}. However, the state space complexity grows exponentially
with state dimensionality. Some works mitigate this by incorporating recurrent neural networks such as RNNs and LSTMs into deep
reinforcement learning frameworks, encoding historical observations to improve performance in POMDPs\cite{b14,b15}. 
Moreover,
\cite{b16} extended state abstraction techniques to historical observation sequences and achieved improved performance in continuous
state-space environments. However, without sufficient exploration, state abstraction tends to degenerate into trivial representations\cite{b11}.

This paper proposes a decoupled hierarchical RL framework integrating state abstraction (DcHRL-SA) for discrete grid environments to address the challenges posed by large-scale exploration spaces and partial observability. The framework designs reasonable high-level goals while employing simple rule-based control for low-level execution. An action mask mechanism is incorporated to restrict invalid actions and improve exploration efficiency\cite{b17}. Furthermore, it is theoretically proven that the decoupled hierarchical mechanism preserves the existence of an optimal policy when the discount factor $\gamma$ is set to 1.
The framework further integrates DeepMDP abstraction to compress the state space and enhance both training efficiency and policy robustness. Experimental results demonstrate that the proposed approach consistently outperforms the baseline PPO algorithm in cumulative rewards and convergence efficiency in MDPs and POMDPs, providing a practical and effective solution for RL in complex discrete grid environments.

\section{Methodology}
\subsection{DcHRL-SA Framework}
\begin{figure*}[htbp]
\centering
\includegraphics[width=0.7\textwidth]{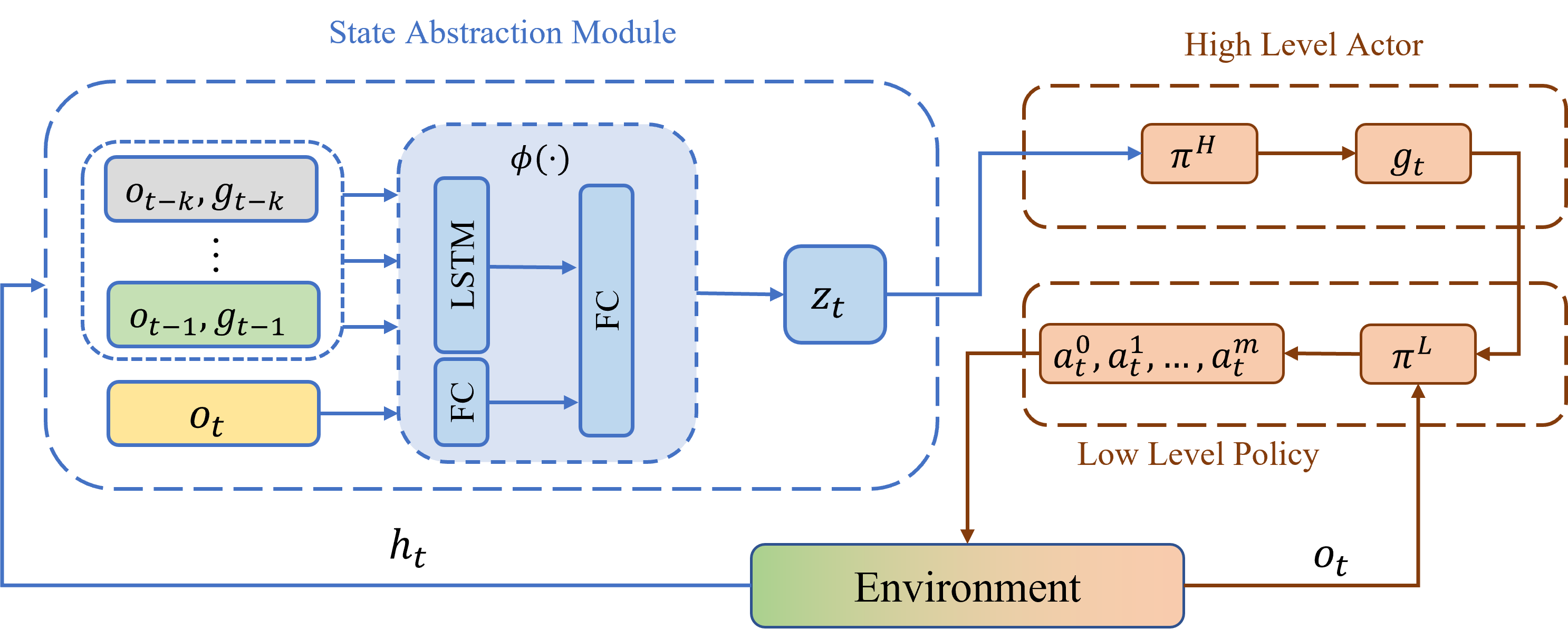}
\caption{Overview of the Combined Framework for Inference in POMDPs.}
\label{decision_process}
\end{figure*}

The decision-making process of the proposed framework in POMDP environments is illustrated in Fig.~\ref{decision_process}.
First, the environment continuously provides a historical observation-goal sequence, denoted as $h_t$, which is formally defined as:
\begin{equation}
h_t = \{ o_{t-l}, g_{t-l}, \dots, o_{t-1}, g_{t-1}, o_t \}
\label{eq:history_goal}
\end{equation}
where $o_i$ and $g_i$ denote the observation and sub-goal at time $i$, respectively, and $l$ is the history length. This sequence is processed by the state abstraction module through the abstraction mapping function $\phi(\cdot)$, producing the latent state representation $z_t$ at the current timestep.
Subsequently, a high-level actor, implemented based on the PPO algorithm, takes $z_t$ as input and generates a new goal $g_t$ for the current decision step. The generated goal $g_t$, together with the current observation $o_t$, is then passed to a rule-based low-level policy, which outputs an action sequence $\{a_t^0, a_t^1, \dots, a_t^m\}$\ required to achieve the goal.
The environment executes this action sequence until completion and transitions to a new observation state. Afterward, the environment provides updated observation information, and the decision-making process proceeds to the next iteration.
In MDP environments, since the agent has full observability, the state $s_t$ is directly passed into the state abstraction function to obtain the abstract state $z_t$, without additional inference steps.
\subsection{DeepMDP Abstraction in POMDPs}
State abstraction methods aim to reduce the dimensionality of state spaces by mapping states with similar dynamic
behaviors into a unified abstract space. The DeepMDP framework leverages end-to-end neural
network models to learn an abstract mapping function $z = \phi(s)$.

A MDP typically consists of a state set $S$, an action set $A$, a state transition function $P:S \times A \times S \rightarrow [0,1]$,
and a reward function $R:S \times A \rightarrow \mathbb{R}$. A POMDP additionally includes an observation space $X$ and an
observation function $O:S \times A \times X \rightarrow [0,1]$, where $O(s', a, x) = P(x \mid a, s')$ denotes the probability
of receiving an observation $x$ after executing action $a$ and transitioning to state $s'$.

Under the assumption of deterministic transitions, where each state-action pair $(s_t, a_t)$ uniquely determines
the next state $s_{t+1} = P(s_t, a_t)$, the DeepMDP constraint is formally expressed as:
\begin{align}
\forall s_1, s_2, \ &\phi(s_1)=\phi(s_2), \ \forall a \in A, , \nonumber \\
&R(s_1, a)=R(s_2, a),\phi(P(s_1, a))=\phi(P(s_2, a))
\label{eq:bisim_condition}
\end{align}

During training, DeepMDP jointly optimizes the abstract mapping, reward prediction, and transition prediction networks by minimizing the following modeling loss:
\begin{align}
L_{\text{bisim}}(\theta_1, \theta_2, \theta_3)
= & \ \mathbb{E}_{s, s', a, r} \bigg[
\left\| R_{\theta_1}(\phi_{\theta_3}(s), a) - r \right\|^2_2 \nonumber \\
& + \lambda \left\| P_{\theta_2}(\phi_{\theta_3}(s), a) - \phi_{\theta_3}(s') \right\|^2_2
\bigg]
\end{align}

An observation-action history sequence for POMDPs is defined as $h_t = \{ o_{t-l}, a_{t-l}, \dots, o_{t-1},a_{t-1}, o_t \}$.
Based on this, we construct an extended state space $\mathcal{H}$.

To effectively capture temporal dependencies in the observation sequences, a LSTM module is incorporated into the abstraction mapping function $\phi(\cdot)$ to encode the dynamic information
within the history sequence through its hidden state.

Building upon this encoding, the DeepMDP constraint is applied to the observation–action histories,
with the corresponding loss function defined as:
\begin{align}
L_{\text{bisim}}(\theta_1, \theta_2, \theta_3)
= & \ \mathbb{E}_{h, h', a, r} \bigg[
\left\| R_{\theta_1}(\phi_{\theta_3}(h), a) - r \right\|^2_2 \nonumber \\
& + \lambda \left\| P_{\theta_2}(\phi_{\theta_3}(h), a) - \phi_{\theta_3}(h') \right\|^2_2
\bigg]
\end{align}

This approach enables the model to extract DeepMDP-consistent structural information from historical
observation sequences while incorporating data augmentation strategies inspired by \cite{b18} to improve the
robustness of the abstraction model. As a result, it enhances the state abstraction capability under partial
observability and improves both policy convergence speed and training stability.

\subsection{Decoupled Hierarchical Strategy}

The proposed DcHRL-SA framework follows the goal-based HRL paradigm proposed in \cite{b6}, where a sub-goal set $G$ is defined to guide hierarchical decision-making.
The high-level actor $\pi^H: S \rightarrow G$ selects a sub-goal $g \in G$ based on the current state $s \in S$,
while the low-level policy $\pi^L: S \times G \rightarrow A$ determines the action $a \in A$ conditioned on both
the current state $s$ and the selected sub-goal $g$. In this framework, sub-goals typically correspond to sub-regions
of the state space $S$ or specific target states. The low-level policy generates action sequences to accomplish the
given sub-goal. The objective of HRL is to maximize the cumulative reward under the joint policy:

\begin{equation}
J(\pi^H, \pi^L) = \mathbb{E}_{\pi^H, \pi^L} \left[ \sum_{t=0}^{\infty} \gamma^t R(s_t, a_t) \right]
\end{equation}

Considering the characteristics of discrete grid environments in terms of observations and action spaces,
this work defines sub-goals as composite actions, which are formed by combining primitive actions from the action set.
This design ensures the stability and reliability of the low-level policy’s execution.
For instance, in a modified \textit{DoorKey} environment, the agent perceives its surroundings through a $\mathcal{W} \times \mathcal{W}$
local observation window centered on itself, where $\mathcal{W}$ represents the size of the observation window.
The basic action space consists of five discrete actions: moving up, down, left, right, and an interaction action.
Composite actions can be categorized into two types: (1) \textit{Move}, which involves a sequence of
movement actions, and (2) \textit{Move and interact}, which consists of a
sequence of movements followed by performing the interaction at the final location.
This work extends the goal space into position–action type pairs, including:

\begin{itemize}
    \item Target position: one of the $\mathcal{W} \times \mathcal{W}$ discrete positions within the local observation.
    \item Action type: either \textit{Move} or \textit{Move and interact}.
\end{itemize}

Thus, the goal space for the high-level actor is expanded to $2\times\mathcal{W} \times \mathcal{W}$ dimensions, specifically defined as:
\begin{itemize}
    \item The first $\mathcal{W} \times \mathcal{W}$ actions: selecting a target position and moving to that location.
    \item The remaining $\mathcal{W} \times \mathcal{W}$ actions: selecting a target position and executing an interaction upon arrival.
\end{itemize}

The low-level policy dynamically generates the optimal movement action sequence from the current position to the specified
target position based on the current observation $o_t$ and the sub-goal provided by the high-level actor,
and performs the corresponding operation if required at the destination.

Formally, the high-level action space is defined as:
\begin{equation}
G = \left\{ g_{i,j,k} \ \big| \ (i,j) \in \mathcal{O},\ k \in \{0,1\} \right\}
\end{equation}
where $(i,j)$ denotes the target position within the observation space $\mathcal{O}$, $k=0$ represents \textit{Move},
and $k=1$ denotes \textit{Move and interact}.

To further enhance decision-making efficiency, this work introduces an action mask mechanism to filter out invalid
sub-goals based on environmental constraints, thereby avoiding ineffective exploration.
Specifically, a validity-checking function $\mathcal{F}(o, g)$ is defined to determine whether a sub-goal $g$
is valid given the current observation state $o$. The validity rules are as follows:
\begin{itemize}
    \item For \textit{Move}: if the target position $(i,j) \in \mathcal{O}$ is unreachable in the current observation, then: $\mathcal{F}(o, g_{i,j,0}) = 0$.
    \item For \textit{Move and interact}: if the target position $(i,j) \in \mathcal{O}$ does not contain any interactive object, then: $\mathcal{F}(o, g_{i,j,1}) = 0$.
\end{itemize}
The set of valid sub-goals is defined as:
\begin{equation}
G_{\text{valid}}(o) = \left\{ g \in G \ \big| \ \mathcal{F}(o, g) = 1 \right\}
\end{equation}

Finally, the agent samples actions only within $G_{\text{valid}}(o)$, thereby improving exploration efficiency and accelerating policy convergence.
Moreover, since the hierarchical goal space fully covers the basic action space,
it can be theoretically shown that the existence of an optimal policy is preserved when the discount factor $\gamma$ is set to 1, as formally proven in Appendix.

\section{Experimental Results}
\begin{figure}[t]
  \centering
  \begin{minipage}[t]{0.49\linewidth}
    \centering
    \includegraphics[width=\linewidth]{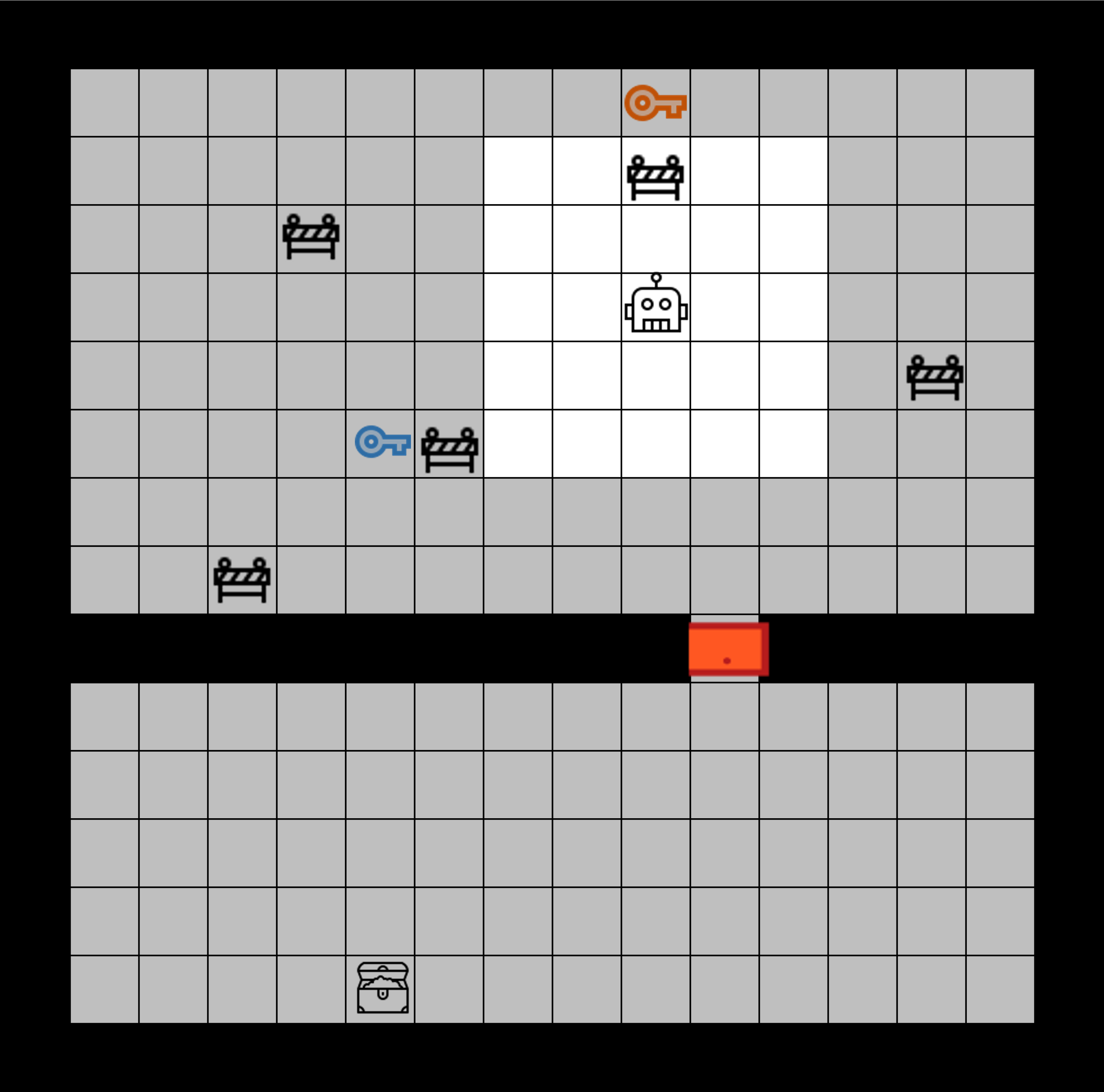}
    \vspace{1mm}
    {\footnotesize(a) Customized DoorKey}
  \end{minipage}
  \hfill
  \begin{minipage}[t]{0.49\linewidth}
    \centering
    \includegraphics[width=\linewidth]{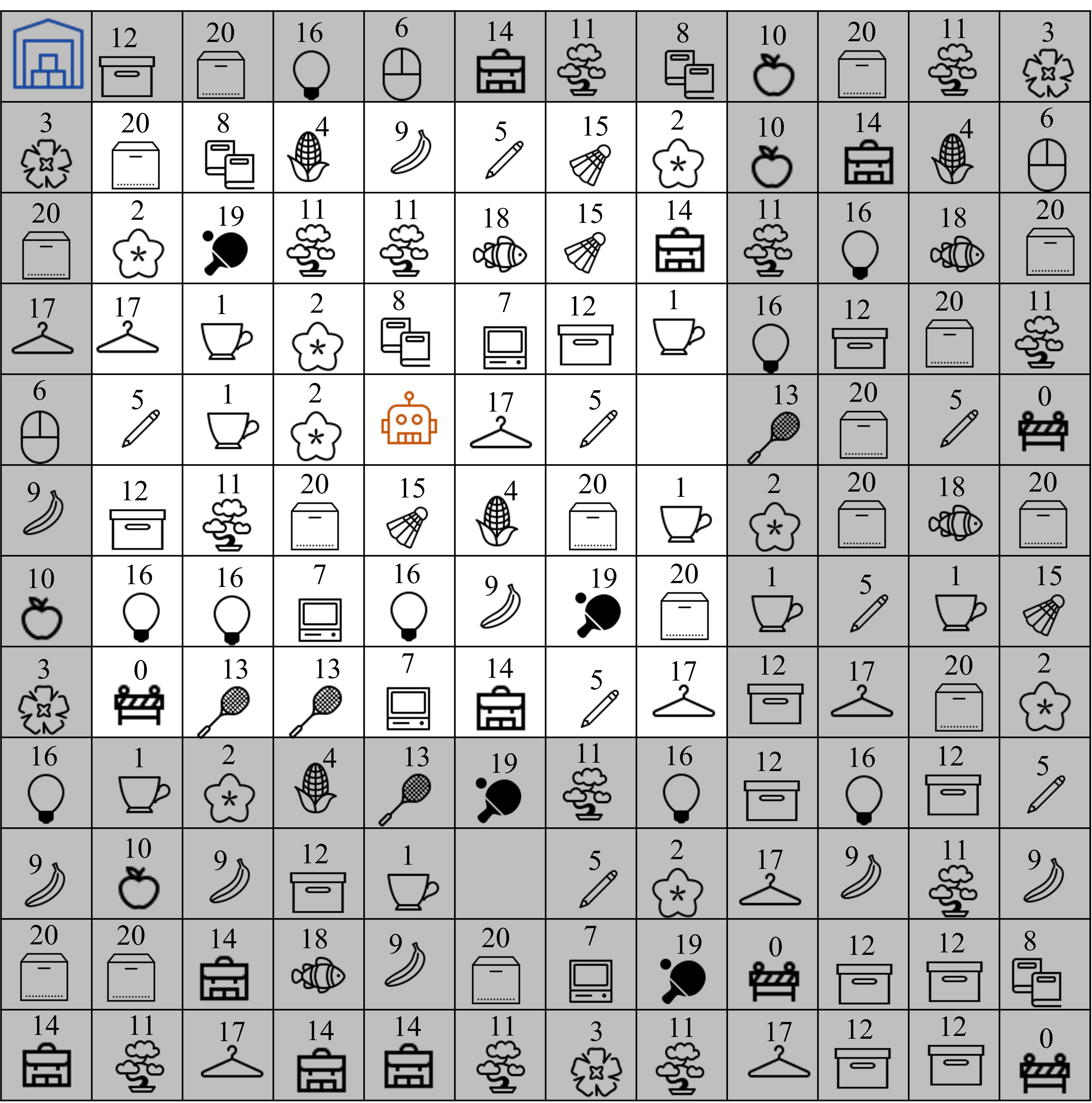}
    \vspace{1mm}
    {\footnotesize(b) Multi-Item collection grid}
  \end{minipage}
  \caption{Visualizations of the discrete grid environments used in experiments.}
  \label{fig:envs_visualization}
\end{figure}

This study conducts experiments on two discrete grid environments to evaluate the performance of the proposed
method under varying task structures and observation conditions. The layouts of two environments are illustrated in Fig.~\ref{fig:envs_visualization}.

\subsection{Environment}
\subsubsection{Customized MiniGrid-DoorKey-16x16}
This environment is based on the DoorKey scenario from the MiniGrid platform \cite{b19}, extended to a $16 \times 16$ grid.
The map is divided by a wall into Room A and Room B. The agent starts in Room A, where two randomly placed keys are available,
but only one can unlock the door to Room B. The agent needs to pick up the correct key, open the door,
and finally reach the treasure chest to receive rewards.

\begin{itemize}
    \item Observation Space: A $5 \times 5$ egocentric view and inventory information.
    \item Action Space: Five discrete actions: \{\textit{up, down, left, right, interact}\}.
    \item Reward Structure:
    \begin{itemize}
        \item +1 for opening the door.
        \item -0.01 penalty for colliding with walls.
        \item $5 - 3 \times (\frac{\text{total\_steps}}{\text{max\_steps}})$ for reaching the treasure chest.
    \end{itemize}
\end{itemize}

\subsubsection{Multi-Item Collection Grid}
This environment is designed as a $12 \times 12$ grid-world with 21 types of randomly distributed items.
There is one warehouse in the position (0,0). The agent collects items into a backpack with unlimited capacity,
and can submit four identical items at the warehouse to gain rewards.

\begin{itemize}
    \item Observation Space:
    \begin{itemize}
        \item \textit{MDP mode:} Full map view and inventory.
        \item \textit{POMDP mode:} A $7 \times 7$ egocentric view and inventory information.
    \end{itemize}
    \item Action Space: Five discrete actions: \{\textit{up, down, left, right, interact}\}.
    \item Reward Structure:
    \begin{itemize}
        \item +3 for successfully submitting a group of four identical items.
        \item $-(0.1+\frac{\text{carry\_number}}{144})$ movement penalty based on the number of items carried.
        \item +100 bonus for completing all submission tasks within the step limit.
        \item $-(140-\text{submit\_number})$ for incomplete tasks.
    \end{itemize}
\end{itemize}
\subsection{Architecture and Training}
Table~\ref{tab:hyperparams} details the key hyperparameter settings adopted in different environments.
These hyperparameters were tuned through preliminary experiments to ensure the stability of the policy
learning process and the effectiveness of state abstraction representations.

\begin{table}[t]
\caption{Key Hyperparameter Settings}
\label{tab:hyperparams}
\centering
\begin{tabular}{c>{\centering\arraybackslash}m{2cm}>{\centering\arraybackslash}m{1.8cm}>{\centering\arraybackslash}m{1.8cm}}
\toprule
\textbf{Module} & \textbf{Hyperparameter} & \textbf{DoorKey-16x16} & \textbf{Multi-Item Grid} \\
\midrule
\multirow{7}{*}{PPO\&LSTM}
    & Optimizer            & Adam    &Adam        \\
    & $lr$   & 0.0001   & 0.0001      \\
    & $\gamma$             & 0.997    & 0.997       \\
    & $\lambda$        & 0.95     & 0.95        \\
    & $\epsilon$  & 0.2      & 0.2         \\
    & $Bs$          & 256      & 256         \\
    & $l$   & 15       & 15          \\
\midrule
\multirow{3}{*}{State Abstraction}
    & $dim(Z)$ & 60 & 25(MDP) 40(POMDP) \\
    & $Bs$                            & 384  & 384  \\
    & $TF$                     & 30   & 30   \\
\midrule
\multirow{1}{*}{Randomness}
    &Environment      & $P_{agent},P_{door}$ $P_{key},P_{wall}$ $P_{treasure}$ & $P_{agent}$\\
\bottomrule
\end{tabular}
\end{table}
The hyperparameters used in the experiments are described as follows. $lr$ denotes the learning rate.
$\gamma$ is the discount factor for future rewards. $\lambda$ represents the coefficient used in Generalized Advantage Estimation (GAE).
$Bs$ indicates the training batch size. $l$ refers to the length of the historical observation sequence employed in POMDPs.
$dim(Z)$ specifies the dimensionality of the abstract state space after state abstraction. $TF$ represents the training frequency per training step.
The parameters $P_{agent}, P_{door}, P_{key}, P_{wall},$ and $P_{treasure}$ denote the initial position of the agent, door, key, wall, and treasure, respectively.
\subsection{Results}
To verify the effectiveness of the proposed method, comparative experiments were conducted in the customized DoorKey environment
and multi-item collection grid environment. The experiments compared three methods: the baseline PPO, decoupled hierarchical RL (DcHRL), and DcHRL-SA. All experiments were performed on an NVIDIA RTX 4090 GPU platform under
consistent hyperparameter settings to ensure fairness.

\begin{figure}[b]
    \centering
    \includegraphics[width=0.45\textwidth]{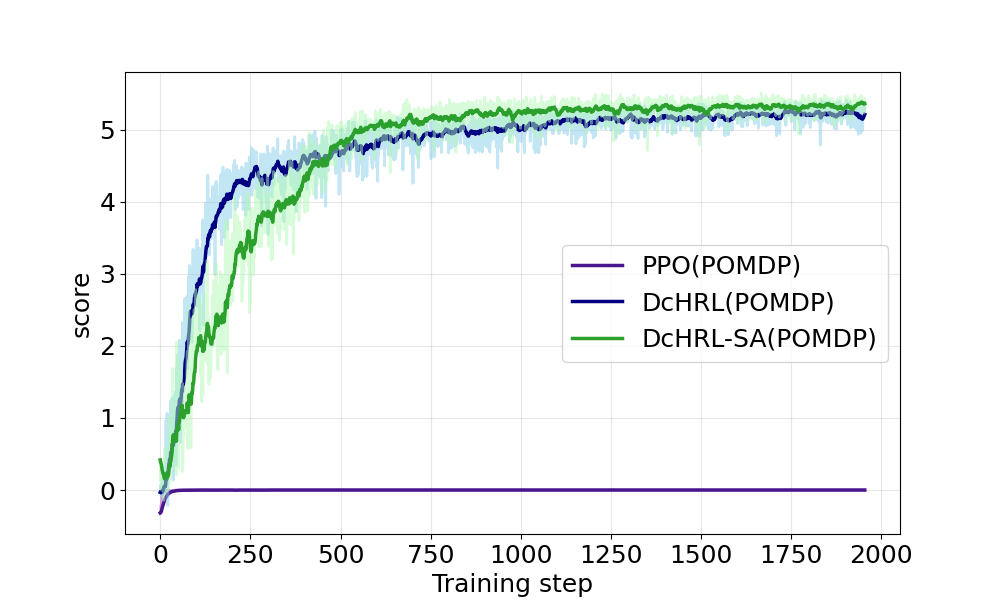}
    \caption{Score curves of different methods in the MiniGrid DoorKey environment.}
    \label{fig:score_curve}
\end{figure}

The score curve for the customized DoorKey environment are illustrated in Fig.~\ref{fig:score_curve}.
Quantitative results are summarized in Table~\ref{tab:quantitative_results}, confirming the advantages of hierarchical decision-making and the potential of state abstraction in discrete grid environments.
Compared with the baseline PPO, the proposed DcHRL framework significantly improves the final score and accelerates policy convergence by leveraging hierarchical exploration strategies.
Moreover, the DcHRL-SA method, which integrates DeepMDP abstraction, slightly improves the final score while effectively reducing the state space dimension from $15 \times 26$ to 60.

\begin{figure}[t]
  \centering
  \begin{minipage}[t]{0.49\linewidth}
    \centering
    \includegraphics[width=\linewidth]{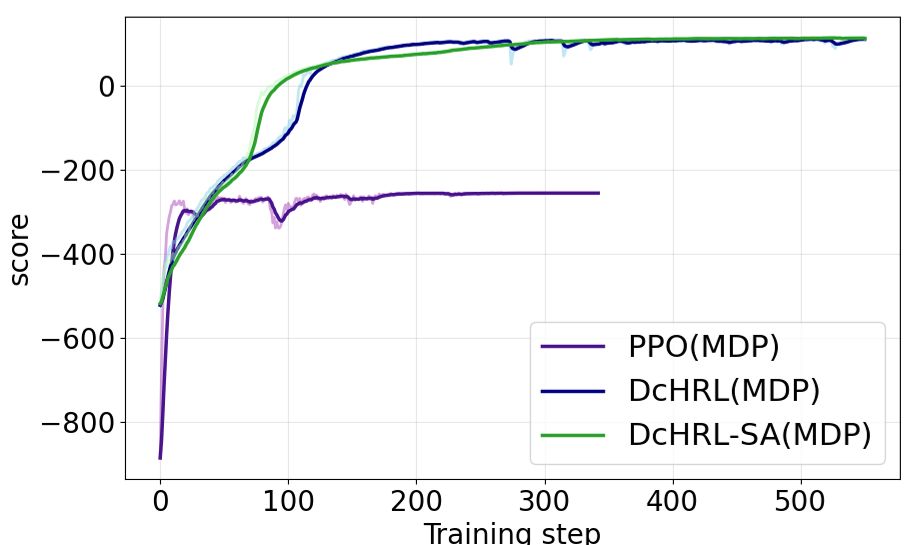}
    \vspace{1mm}
    {\footnotesize(a) Score in MDP}
  \end{minipage}
  \hfill
  \begin{minipage}[t]{0.49\linewidth}
    \centering
    \includegraphics[width=\linewidth]{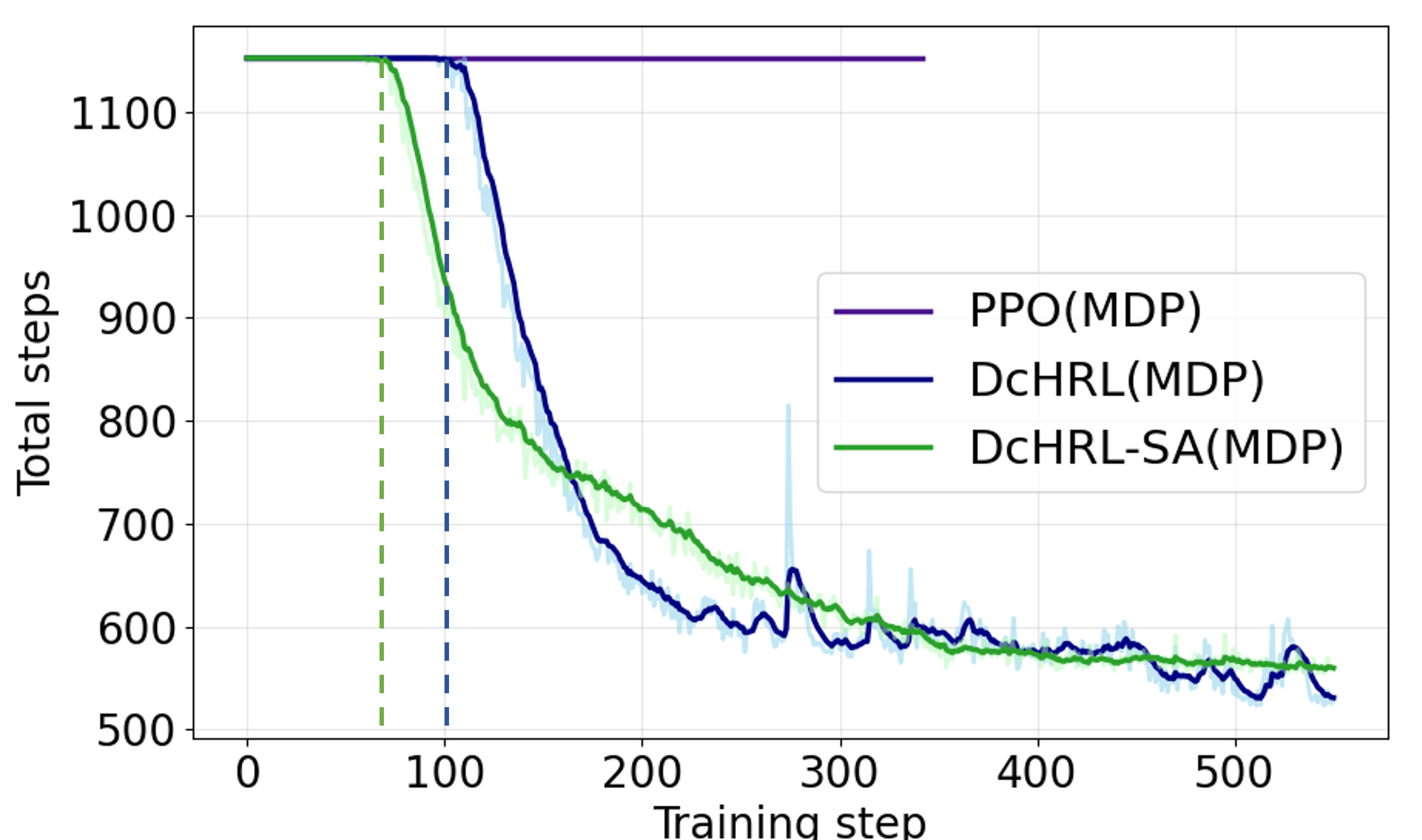}
    \vspace{1mm}
    {\footnotesize(b) Steps in MDP}
  \end{minipage}

  \vspace{1ex} 

  \begin{minipage}[t]{0.49\linewidth}
    \centering
    \includegraphics[width=\linewidth]{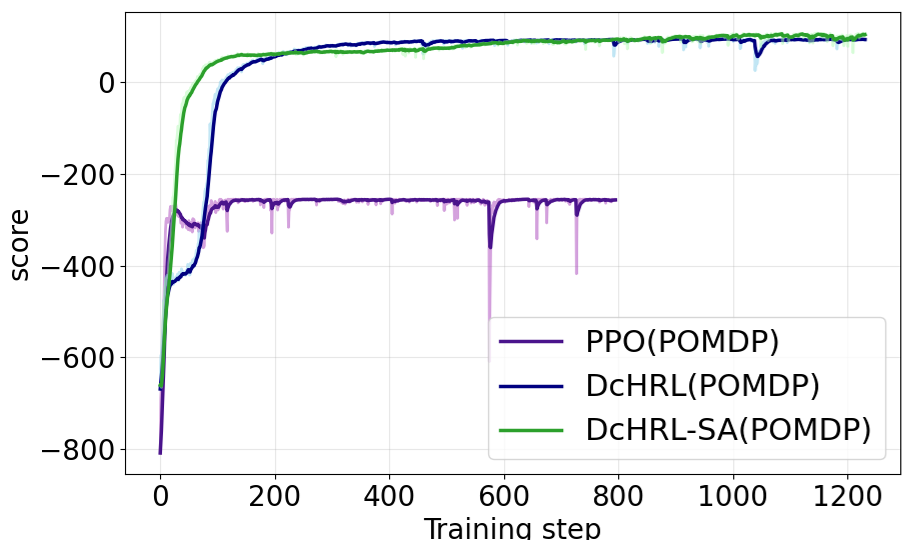}
    \vspace{1mm}
    {\footnotesize(c) Score in POMDP}
  \end{minipage}
  \hfill
  \begin{minipage}[t]{0.49\linewidth}
    \centering
    \includegraphics[width=\linewidth]{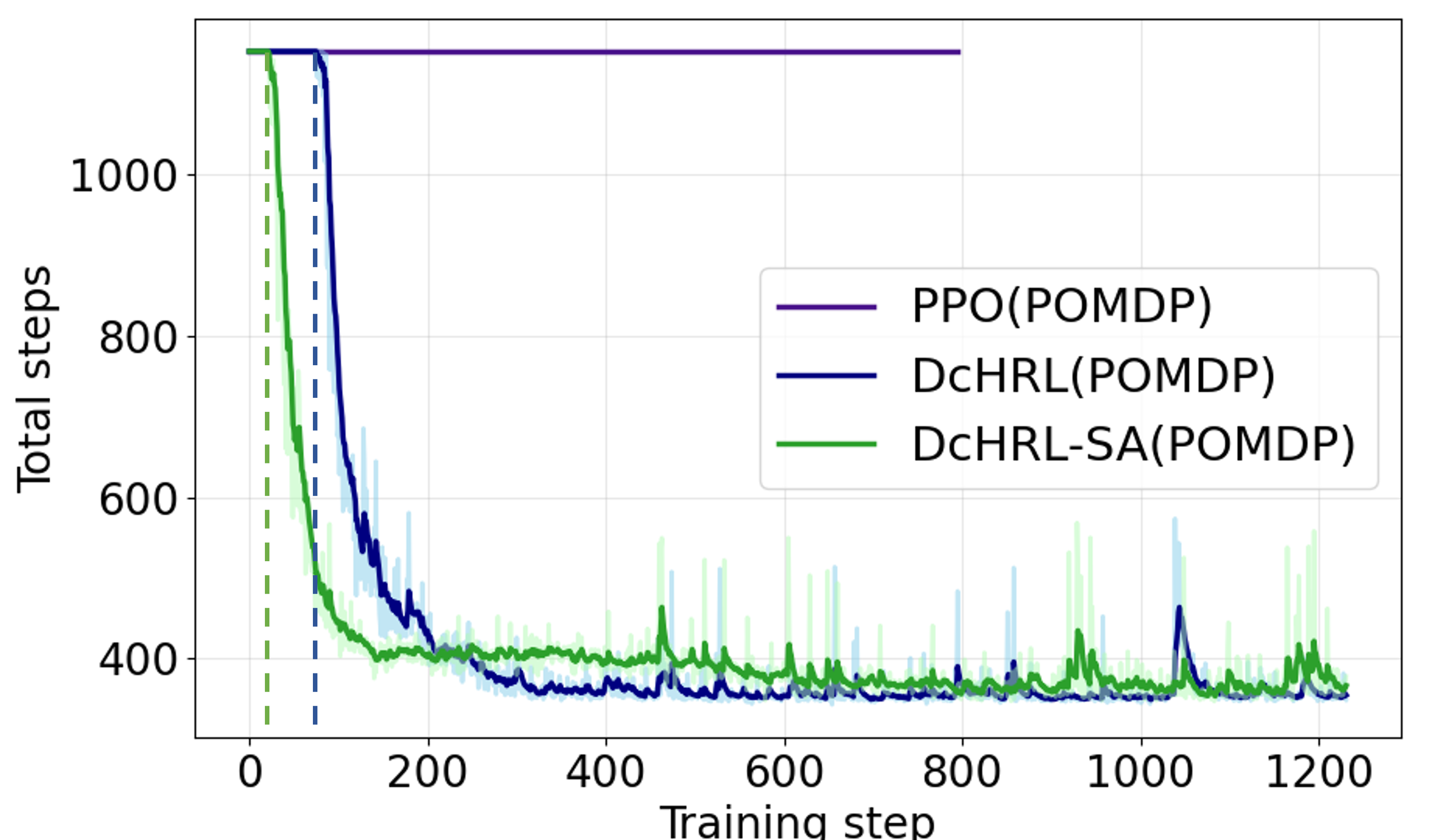}
    \vspace{1mm}
    {\footnotesize(d) Steps in POMDP}
  \end{minipage}

  \caption{Training performance under different methods in MDP and POMDP environments.}
  \label{fig:custom_env_curves}
\end{figure}

In the multi-item collection grid environment, similar performance trends are observed. The score curves and task completion step curves for both MDP and POMDP modes are illustrated in Fig.~\ref{fig:custom_env_curves}, 
and the corresponding quantitative results are summarized in Table~\ref{tab:custom_env_results}.
In terms of score performance, the baseline PPO algorithm demonstrates poor exploration capability in both environments, quickly converging to local optima. In contrast, the DcHRL framework substantially improves exploration efficiency and final performance in both MDP and POMDP modes. Furthermore, the DcHRL-SA framework achieves additional gains in final converged scores while effectively reducing the state dimensionality.

Regarding the task completion steps, the step curves of DcHRL-SA start to decline earlier than those of DcHRL as additionally marked in Fig.~\ref{fig:custom_env_curves}. DcHRL-SA exhibits smoother and more stable convergence curves than DcHRL in the MDP mode, indicating enhanced optimization stability. In the POMDP mode, while DcHRL-SA converges slightly faster than DcHRL, its step curves show increased oscillation near convergence, where the incompleteness of information makes it difficult to converge to a stable abstraction function.

Overall, these results validate the effectiveness of DcHRL-SA framework, enhancing both exploration efficiency and convergence speed in discrete grid environments under varying observability conditions.
\begin{table}[t]
\caption{Performance Comparison in MiniGrid DoorKey Environment}
\label{tab:quantitative_results}
\centering
\begin{tabular}{>{\centering\arraybackslash}m{3cm}>{\centering\arraybackslash}m{2cm}>{\centering\arraybackslash}m{2cm}}
\toprule
\textbf{Method} & \textbf{Final Reward} & \textbf{State Dimension} \\
\midrule
PPO & $0.0000 \pm 0.0000$ & $15 \times 26$ \\
DcHRL & $5.2158 \pm 0.0981$ & $15 \times 26$ \\
DcHRL-SA & $5.3207 \pm 0.0849$ & 60 \\
\bottomrule
\end{tabular}
\end{table}
\begin{table}[t]
\caption{Performance Comparison in Custom Grid Environments (MDP \& POMDP Modes)}
\label{tab:custom_env_results}
\centering
\begin{tabular}{cccc}
\toprule
\textbf{Modes} &\textbf{Method} & \textbf{Converged Score} &\textbf{Total steps} \\
\midrule
\multirow{3}{*}{MDP} &PPO & $-263.81 \pm 0.00$ & $1152$ \\
                            & DcHRL & $108.40 \pm 3.41$ & $547.62 \pm 14.38$ \\
                    & DcHRL-SA & $114.10 \pm 0.80$ & $561.87 \pm 4.74$ \\
\midrule
\multirow{3}{*}{POMDP} &PPO & $-267.18 \pm 33.27$ & $1152$ \\
                            & DcHRL & $90.33 \pm 5.49$ & $363.13 \pm 16.96$ \\
                    & DcHRL-SA & $100.95 \pm 4.61$ & $369.70 \pm 24.17$ \\
\bottomrule
\end{tabular}
\end{table}

\section{Conclusion}
This paper proposes a decoupled hierarchical reinforcement learning framework combined with a DeepMDP abstraction method,
aiming to improve decision-making efficiency and exploration capability in sparse-reward discrete grid environments.
Experimental results on the customized DoorKey environment and a multi-item collection grid environment demonstrate that
the decoupled hierarchical policy outperforms the PPO baseline in both convergence speed and final performance,
validating the effectiveness of hierarchical decision structures in improving exploration in complex discrete environments.
Although the introduction of the DeepMDP abstraction mechanism results in limited improvement in final scores,
experiments show that it successfully captures essential decision-related information and exhibits a clear advantage in convergence
speed in both MDPs and POMDPs. This confirms the potential of state abstraction methods in enhancing training efficiency
and model scalability. Overall, this work provides practical guidance and methodological references for future research on optimizing hierarchical policies and state abstraction techniques.

\appendix
\setcounter{equation}{0}
\renewcommand{\theequation}{A.\arabic{equation}}
\begin{proof}
The notations used in this proof are summarized as follows:
\begin{itemize}
    \item $\mathcal{S} = \{s_1, s_2, \ldots, s_n\}$: the finite set of $n$ discrete environment states.
    \item $\mathcal{A} = \{a_1, a_2, \ldots, a_m\}$: the finite set of $m$ possible actions.
    \item $\mathbf{u}^{\pi}_t \in \mathbb{R}^{n \times 1}$: the state distribution vector at time $t$ under policy $\pi$, where the $i$-th entry $u^{\pi}_t(s_i)$ represents the probability of being in state $s_i$ at time $t$.
    \item $\mathbf{P}^{\pi} \in \mathbb{R}^{n \times n}$: the state transition probability matrix under policy $\pi$, where the $(i,j)$-th entry $p^{\pi}(s_i|s_j)$ represents the probability of transitioning from state $s_j$ to state $s_i$.
    \item $\pi(a|s)$: the probability of selecting action $a$ when in state $s$ under policy $\pi$.
    \item $R(s, a)$: the immediate reward obtained by taking action $a$ in state $s$.
    \item $\mathbf{R}^{\pi} \in \mathbb{R}^{n \times 1}$: the expected immediate reward vector under policy $\pi$.
    \item $\mathbf{V}^{\pi} \in \mathbb{R}^{n \times 1}$: the state value function vector under policy $\pi$, where the $i$-th entry $V^{\pi}(s_i)$ represents the expected cumulative discounted reward starting from state $s_i$.
    \item $\gamma \in [0, 1]$: the discount factor for future rewards.
    \item $J^{\pi}(\mathbf{u}^{\pi}_t)$: the expected cumulative reward at time $t$ under policy $\pi$.
\end{itemize}

Accordingly, for a given policy $\pi$, the corresponding state probability transition trajectory can be denoted as $\mathbf{u}_0 \rightarrow \mathbf{u}^{\pi}_1 \rightarrow \mathbf{u}^{\pi}_2 \rightarrow \cdots \rightarrow \mathbf{u}_e$, where $\mathbf{u}_e$ represents the desired target distribution, which is independent of the policy $\pi$. 

If the state distributions at two time steps along the trajectory, $\mathbf{u}^{\pi}_i$ and $\mathbf{u}^{\pi}_j$ $(i \leq j)$, are known, the expected cumulative reward when transitioning from $\mathbf{u}^{\pi}_i$ to $\mathbf{u}^{\pi}_j$ under policy $\pi$ is given by:
\begin{equation}
d_{\pi}(\mathbf{u}^{\pi}_i, \mathbf{u}^{\pi}_j) = \sum_{t=i}^{j-1} \gamma^{t-i} \left(\mathbf{u}^{\pi}_t\right)^T \mathbf{R}^{\pi}
\label{eq:expected-reward}
\end{equation}
which depends only on the policy $\pi$ and the two state distributions $\mathbf{u}^{\pi}_i$ and $\mathbf{u}^{\pi}_j$.

The optimal policy $\pi^*$ is then defined as:
\begin{equation}
\pi^* = \arg\max_{\pi} d_{\pi}(\mathbf{u}_0, \mathbf{u}_e)
\end{equation}

Notably, given an intermediate point $\mathbf{u}_m$ on the optimal trajectory, it holds that:
\begin{equation}
d_{\pi^*}(\mathbf{u}_0, \mathbf{u}_e) = d_{\pi^*}(\mathbf{u}_0, \mathbf{u}_m) + \gamma^m d_{\pi^*}(\mathbf{u}_m, \mathbf{u}_e)
\end{equation}

Particularly when $\gamma=1$, it simplifies to:
\begin{equation}
d_{\pi^*}(\mathbf{u}_0, \mathbf{u}_e) = d_{\pi^*}(\mathbf{u}_0, \mathbf{u}_m) + d_{\pi^*}(\mathbf{u}_m, \mathbf{u}_e)
\end{equation}

Therefore, the optimal policy $\pi^*$ is also the optimal policy for both sub-problems $\mathbf{u}_0 \rightarrow \mathbf{u}_m$ and $\mathbf{u}_m \rightarrow \mathbf{u}_e$ independently.

When the precise intermediate point on the optimal path is not known, but only an abstract set $G$ containing possible intermediate states $\mathbf{g}$, the problem can be expressed as:

\begin{equation}
\begin{split}
\max_{\pi} d_{\pi}(\mathbf{u}_0, \mathbf{u}_e) 
= \ & \max_{\pi_1, \pi_2, \mathbf{g}} \Big( d_{\pi_1}(\mathbf{u}_0, \mathbf{g}) 
+ d_{\pi_2}(\mathbf{g}, \mathbf{u}_e) \Big) \\
= \ & \max_{\mathbf{g}_1} \Big( \max_{\pi_1} d_{\pi_1}(\mathbf{u}_0, \mathbf{g}_1) 
+ \max_{\mathbf{g}_2} \big( \\
& \max_{\pi_2} d_{\pi_2}(\mathbf{g}_1, \mathbf{g}_2) 
+ \max_{\pi_3} d_{\pi_3}(\mathbf{g}_2, \mathbf{u}_e) \big) \Big)
\end{split}
\label{eq:split-opt}
\end{equation}

When $\mathbf{g}_1$ equals to an optimal point $u^*_1$ on the optimal path:
\begin{equation}
\begin{split}
\max_{\pi} d_{\pi}(\mathbf{u}_0, \mathbf{u}_e) 
= \max_{\pi_1} d_{\pi_1}(\mathbf{u}_0, \mathbf{u}_1^*)+ \\
\max_{\mathbf{g}_2} \Big( \max_{\pi_2} d_{\pi_2}(\mathbf{u}_1^*, \mathbf{g}_2)
+ \max_{\pi_3} d_{\pi_3}(\mathbf{g}_2, \mathbf{u}_e) \Big)
\end{split}
\label{eq:split-step}
\end{equation}

Recursively, when each $\mathbf{g}_i$ equals to the corresponding optimal point $u^*_i$, we obtain:
\begin{equation}
\begin{aligned}
d_{\pi^*}(\mathbf{u}_0, \mathbf{u}_e) = 
&\ \max_{\pi_1} d_{\pi_1}(\mathbf{u}_0, \mathbf{u}_1^*) 
+ \max_{\pi_2} d_{\pi_2}(\mathbf{u}_1^*, \mathbf{u}_2^*) \\
&\ + \cdots 
+ \max_{\pi_m} d_{\pi_m}(\mathbf{u}_m^*, \mathbf{u}_e)
\end{aligned}
\label{eq:multi-step}
\end{equation}

Finally, since the proposed method's goal space fully covers the base action space, it can degenerate into a flat decision-making process under worst-case conditions, thereby ensuring policy optimality.
\end{proof}
\end{document}